\documentclass{article}

\usepackage{microtype}
\usepackage{graphicx}
\usepackage{subfigure}
\usepackage{placeins}
\usepackage{booktabs} 

\usepackage{hyperref}

\usepackage[accepted]{icml2020}

\icmltitlerunning{Detecting Hate Speech in multi-modal Memes}

\begin{document}

\twocolumn[
\icmltitle{Detecting Hate Speech in multi-modal Memes}



\icmlsetsymbol{equal}{*}

\begin{icmlauthorlist}
\icmlauthor{Abhishek Das}{equal,CMU}
\icmlauthor{Japsimar Singh Wahi}{equal,CMU}
\icmlauthor{Siyao Li}{equal,CMU}

\end{icmlauthorlist}

\icmlaffiliation{CMU}{Carnegie Mellon University, Pittsburgh, United States}
\icmlaffiliation{CMU}{Carnegie Mellon University, Pittsburgh, United States}
\icmlaffiliation{CMU}{Carnegie Mellon University, Pittsburgh, United States}



\vskip 0.3in
]





\section{Abstract}
\label{submission}
In the past few years, there has been a surge of interest in multi-modal problems, from image captioning to visual question answering and beyond. In this paper, we focus on hate speech detection in multi-modal memes wherein memes pose an interesting multi-modal fusion problem. We try to solve the Facebook Meme Challenge \cite{kiela2020hateful} which aims to solve a binary classification problem of predicting whether a meme is hateful or not. A crucial characteristic of the challenge is that it includes "benign confounders" to counter the possibility of models exploiting unimodal priors. The challenge states that the state-of-the-art models perform poorly compared to humans. During the analysis of the dataset, we realized that majority of the data points which are originally hateful are turned into benign just be describing the image of the meme. Also, majority of the multi-modal baselines give more preference to the hate speech (language modality). To tackle these problems, we explore the visual modality using object detection and image captioning models to fetch the “actual caption” and then combine it with the multi-modal representation to perform binary classification. This approach tackles the benign text confounders present in the dataset to improve the performance. Another approach we experiment with is to improve the prediction with sentiment analysis. Instead of only using multi-modal representations obtained from pre-trained neural networks, we also include the unimodal sentiment to enrich the features. We perform a detailed analysis of the above two approaches, providing compelling reasons in favor of the methodologies used.


\begin{figure}
\vskip 0.2in
\begin{center}
\centerline{\includegraphics[width=\columnwidth,height= 5 cm]{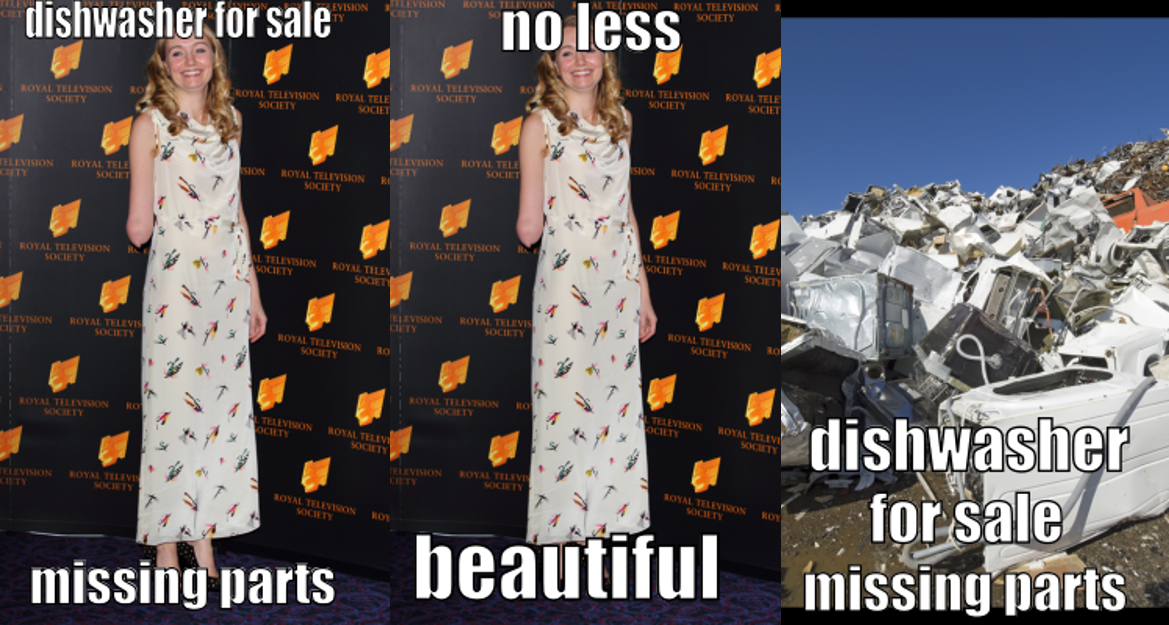}}
\caption{Multi-modal “mean” meme and Benign confounders. Mean meme (left), Benign text confounder (middle) and Benign image confounder (right)}
\label{confounders}
\end{center}
\vskip -0.4in
\end{figure}

\section{Introduction}
\label{submission}

In today's world, social media platforms play a major role in influencing people’s everyday life. Though having numerous benefits, it also has the capability of shaping public opinion and religious beliefs across the world. It can be used to attack people directly or indirectly based on race, caste, immigration status, religion, ethnicity, nationality, sex, gender identity, sexual orientation, and disability or disease. Hate Speech on online social media can trigger social polarization, hateful crimes. On large platforms such as Facebook and Twitter, it becomes practically impossible for a human to monitor the source and spreading of such malicious activities, thus it is the responsibility of the machine learning and artificial intelligence research community to address and solve this problem of detecting hate speech efficiently. 

In tasks such as VQA and multi-modal machine translation, it has been observed that baseline models using the language domain perform well without even exploiting the multi-modal understanding and reasoning\cite{devlin2015exploring}.  However, the Facebook Hateful Memes Challenge dataset is designed in such a manner that unimodal models exploiting just the language or vision modalities separately will fail, and only the models that can learn the true multi-modal reasoning and understanding will be able to perform well. 

They achieve this by the introduction of “benign confounders” in the dataset, i.e. for every hateful meme, they find an alternative image or caption which when replaced, is enough to make the meme harmless or non-hateful, thus flipping the label. Consider a sentence like “dishwasher for sale, missing parts”. Unimodally, this sentence is harmless, but when combined with an equally harmless image of a girl without a hand, suddenly it becomes mean. See Figure \ref{confounders} for an illustration. Thus, this challenge set is an excellent stage that aims to facilitate the development of robust multi-modal models, and at the same time addresses an important real-world problem of detecting hateful speech on online social media platforms. {Majority of the prior work baselines aim at solving this problem by finding an alignment between the two modalities, but it faces the hardship of not knowing the context behind the image and the text combination.}

In this paper, we introduce two major ideas wherein we try to explore the two modalities using pre-trained Image captioning models and sentiment analysis to understand the context and relationship between the two modalities. Many of the baselines tend to focus more on the text modality for hate speech. Also, during the data analysis, we realized that majority of the hateful memes are converted into benign just by describing the image, i.e., benign text confounders. In our first approach, we try to balance the representations of the two modalities and tackle the benign text confounders by fetching a deeper understanding of the image via object detection and captioning. We then use this representation and fuse it with the multi-modal representation from the state-of-the-art models to improve the performance. Through the error analysis, we also found that the finetuning a model with pretrained multi-modal representations does not always provide desirable results. It may because those embeddings are pretrained to predict the semantic correlation between image and text but semantic information are difficult to capture and may be insufficient for solving this challenge. Thus, we try to include some high-level features like text and image sentiments to aid the prediction because the sentiment analysis is a related and relatively simple task. On the Facebook Hateful Memes Challenge Dataset, we show both our approaches benefit the prediction.


In what follows, we discuss the related prior work for such a problem in the next section (3), followed by defining the problem statement (4) and discussing our novel approaches in section (5). We then present our Experimental setup in section (6) followed with its results and discussion in section (7). Finally, we end the discussion with conclusion and future directions in the last section (8).

\section{Related Work}

Hate speech detection has gained more and more attentions in recent years. There have been several text-only hate speech datasets released, mostly based on Twitter [\cite{waseem-2016-racist},\cite{waseem-hovy-2016-hateful},\cite{DBLP:journals/corr/DavidsonWMW17}], and various architectures have been proposed for classifiers [\cite{kumar-etal-2018-benchmarking}, \cite{malmasi-zampieri-2017-detecting}]. Also, in the past few years, there has been a surge in multi-modal tasks and problems, ranging from visual question answering[\cite{Goyal_2017_CVPR}] to image captioning[\cite{sidorov2020textcaps},\cite{gurari2020captioning}] and beyond. However, there has been surprisingly little work related to multi-modal hate speech, with only a few papers including both image and text modality. Some of the works related to multi-modal hate detection based on image and text modality are as follows.

Yang et al. [\cite{yang-etal-2019-exploring}] reported that augmenting text with image embedding information immediately boosts performance in hate speech detection. In this paper, the image embeddings are formed by using the second last layer of the pre-trained ResNet neural network on ImageNet and then hashing these values for efficient photo indexing, searching, and clustering. The most straightforward way of integrating text with photo features is to concatenate both image and text vectors. The concatenated vector is followed by dropout, MLP, and softmax operations for the final hate speech classification. They also explore other fusion techniques like gated summation and bi-linear transformation.


\begin{figure}
\vskip 0.2in
\begin{center}
\centerline{\includegraphics[width=\columnwidth,height= 4 cm]{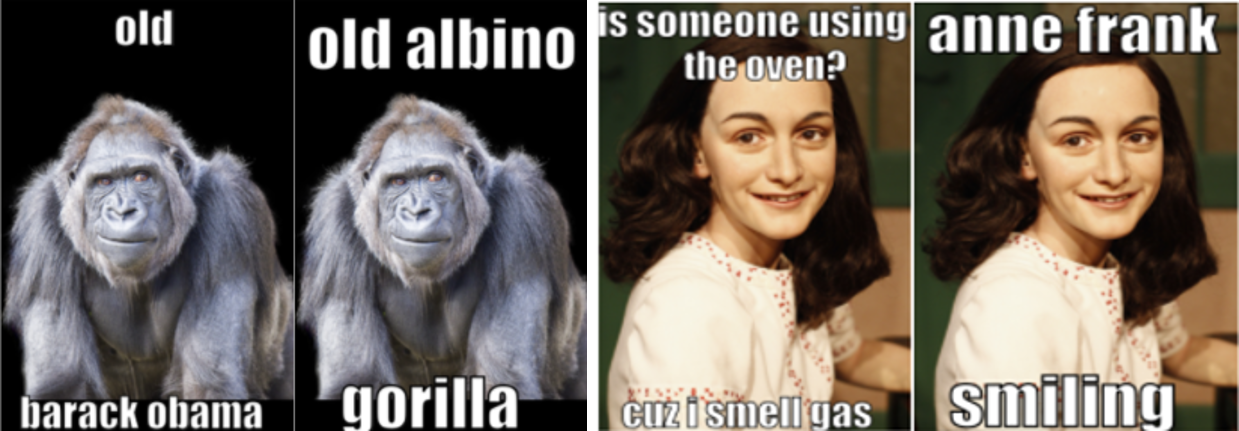}}
\caption{Mean memes and their benign text confounders}
\label{Benign Text Confounders}
\end{center}
\vskip -0.4in
\end{figure}

\begin{figure*}[t]
  \includegraphics[width=\textwidth]{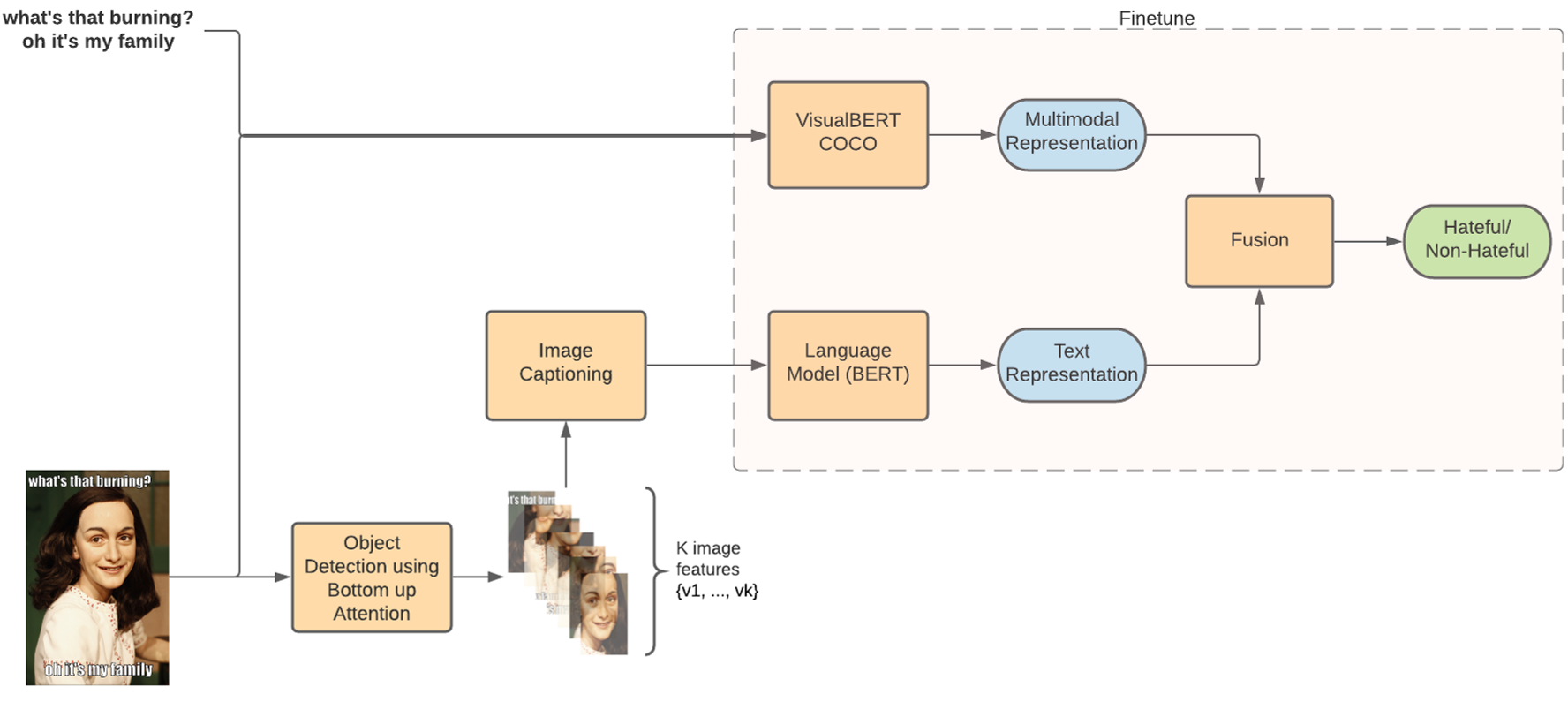}
  \caption{Approach 1 - Model architecture for Image captioning}
\label{approach1}
\end{figure*}

Gomez et al.\cite{Gomez_2020_WACV} highlighted the issue that most of the previous work on hate speech is done using textual data only and that hate-speech detection on multi-modal publications has not been addressed yet. So, they created MMHS150k, a manually annotated multi-modal hate speech dataset formed by 150,000 tweets, each one of them containing text and an image. The data points are labeled into one of the six categories: No attacks to any community, racist, sexist, homophobic, religion-based attacks, or attacks to other communities. They trained a LSTM model which considered just the tweets text as a baseline for the task of detecting hate speech in multi-modal publications. Their further objective was to exploit the information in the visual domain to outperform their baseline. They did this by proposing two models. The first one was the Feature Concatenation Model (FCM), which is an MLP that concatenates the image representation extracted by a CNN and the textual features of both the tweet text and the image text extracted by an LSTM. Their second model named Textual Kernels Model (TKM) was inspired by VQA tasks and was based on the intuition of looking for patterns in the image corresponding to the associated texts. This was done by learning kernels from textual representations and convolving them with CNN feature maps.

Our first approach extends this idea of a deeper understanding of the visual domain. To our knowledge, this paper is the first to use pre-trained image captioning models to generate the "actual caption" from the image along with the image embeddings and add these through fusion techniques like concatenation and bilinear transformations with the multi-modal embedding of the state-of-the-art baselines. 

Now, we describe some relevant work in Image Captioning. \cite{xu2015show} introduced an encoder-decoder architecture which uses attention mechanism to generate captions. It is trainable by standard back-propagation methods. Most conventional approaches use a top-down mechanism for captioning tasks. A recent method \cite{anderson2018bottom} combines Bottom-Up and Top-Down Attention which utilizes a Faster R-CNN based object detection to extract k image features, $V = \{v_1, ..., v_k\}, v_i \in R^{D}$ that enables the attention to be calculated at the level of objects. Each image feature here encodes a salient image region. The captioning model uses a soft-top down approach given the features and partial output sequences as context. It consists of a 2-Layer LSTM, the first layer is called as Top-Down Attention LSTM, the output of which is used to find the attention weights. These attended image features are used by the second LTSM layer which is characterized as a Language Model. They further use cross-entropy loss minimization. The quality of captions generated is vastly improved using this combined technique. Their method is highly modular and allows using various architectures in captioning stage for the features generated using object detection. One can also use different object detection mechanisms in place of Faster R-CNN, or even replace it with the spatial output of a CNN.


Multi-modal sentiment analysis is a relatively new topic. However, extensive research \cite{SOLEYMANI20173} \cite{shenoy2020multilogue} \cite{kumar2020gated} \cite{ghosal2018contextual} \cite{zadeh2018multimodal} \cite{majumder2018multimodal} has already been done in this field and yielded fruitful results. Some \cite{kumar2020gated} \cite{ghosal2018contextual} tends to improve the prediction accuracy by developing more sophisticated attention mechanism to better capture the interaction between two modalities, while some \cite{zadeh2018multimodal} \cite{majumder2018multimodal} introduce very innovative ways of fusion, which utilizes graph or hierarchical architecture. In addition, \cite{shenoy2020multilogue} leverages the sentiments to improve multi-modal dialogue task. However, very little has been done to improve the hateful media detection with multi-modal sentiment information. We introduce a sentiment analysis approach as our second experiment wherein we carry out uni-modal sentiment analysis on both text and visual domains to find the orientation of both the modalities.

\begin{figure*}[h]
  \includegraphics[width=\textwidth]{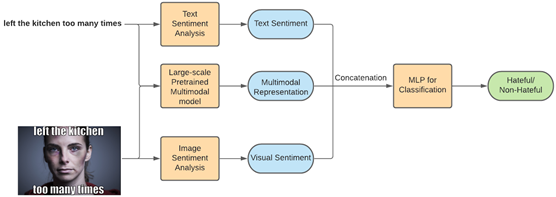}
  \caption{Approach 2 - Model architecture using Sentiment analysis}
\label{approach2}
\end{figure*}

\section{Proposed Approaches}

\subsection{Problem Statement}
The objective of this challenge is to classify memes as hateful or benign while considering their information from both text and visual modality. Denote the visual components of all memes by $X_1 = \{I_1, ..., I_i\}$ where $i$ is the index of the memes, and in our case, the visual component $I$ is the meme itself. Let $X_2 = \{T_1,...,T_i\}$ denotes the text extracted from the memes. If phrases locate in multiple regions of a single meme, the corresponding $T$ will include all the text information by concatenation. Let $Y = \{y_1,...,y_i\}$ be the corresponding labels of all memes, where each $y \in \{0, 1\}$ with 0 means benign and 1 indicates a hateful meme. Thus, our task can be formulated as a binary classification problem with $X_1$ and $X_2$ as input. The goal of our paper is to model the $P(Y|X_1, X_2)$, denoted by $p_\theta$, which minimize the following cost function: 
\begin{equation}
J(\theta) = \sum_i - ( Y log(p_\theta) + (1 - Y) log (1 - p_\theta))
\end{equation}

\subsection{Image Captioning}
As discussed above, this paper tackles the benign text confounders present in the dataset which converts an originally hateful meme into a benign one just by describing what is happening in the image. Figure \ref{Benign Text Confounders} shows some of these adversarial samples. As shown in Figure \ref{type_of_memes}, they account for 20\% of the dataset and thus our hypothesis is that if we can provide our model with this extra knowledge, it will combat these adversarial examples and provide a boost in accuracy. Using object detection and image captioning helps in learning this aspect of the dataset and understanding the behavior of the benign text confounders and thus gives a better performance than the baseline models. Comparing the "actual caption" with the "pre-extracted caption" of the meme will help in understanding whether both are aligned or not. Also, most of the multi-modal baselines tend to focus more on the text modality for the hate speech. Our intuition behind this approach is to find a deeper relationship between the text and the image modalities. 

As we can see in the Figure \ref{approach1}, we fist pass the hateful dataset (both the modalities,i.e., X- pre-extracted captions and Y- Images of the meme) into the Visual Bert model pre-trained on the COCO dataset. This fetches us the multi-modal representation of the two modalities, i,e., a 786 tensor of the multi-modal representation (m1,m2,m3,...). Parallelly, we also pass the image into an Image Captioning model (Show, Attend, and Tell, Bottom up top down), which fetches us a caption for the image present in the meme ($X_3 = \{C_1,...,C_i\}$ denotes the caption extracted from the images.). We then pass this text caption via a pre-trained Bert model to get a textual representation of another 768 dimensional tensor (h1,h2,h3,...). Then, we fuse the two tensors using fusion techniques like concatenation and bilinear transformations. Bilinear transformation is a filter to integrate the information of two vectors into one vector. Mathematically we have bilinear (m',h', dim) = m'$^T$.M.h + b, where dim is a hyper-parameter indicating the expected dimension of the output vector (768), M is a weight matrix of dimension (dim,$|$m'$|$,$|$h'$|$), and b is a bias vector of dimension dim. Again we concatenate m, h, and bilinear(m',h',dim) for hate speech classification. Finally, we then pass the output via an Multi-layer perceptron to get a binary classification of hateful and non-hateful memes (0/1).

We fine tune the Visual Bert model and the Bert model from the Facebook hateful dataset and the captions generated on the images of the Facebook hateful dataset. This new approach of combining the image captioning and multi-modal baselines helps in tackling the previous mentioned challenges and increases the performance significantly.

\subsection{Sentiment Analysis}
Another approach is to utilize the sentiment information of both modalities to generate richer representations for further prediction. We first obtain the multi-modal contextual representations $e_m$ from input $T$ and $I$ by using a pre-trained model. In our experiment, we use VisualBERT \cite{li2019visualbert}. However, similar to some other pre-trained models, the VisualBERT focuses more on the correlation between the input modalities, but the text and image in hateful memes are usually connected indirectly. Thus, unimodal sentiments, which are closely related to hate detection, can benefit the prediction. A RoBERTa \cite{liu2019roberta} model is then used to obtain the text sentiment embeddings $e_t$ from $T$, while a VGG \cite{simonyan2014very} is applied for visual sentiments $e_v$ from $I$. However, due to the limitation of annotated data, we are unable to fine-tune those two models on our dataset. Instead, the RoBERTa is trained on Stanford Sentiment Treebank \cite{socher2013recursive} and the visual sentiment model parameters are learned from T4SA dataset \cite{vadicamo2017cross}. Then, $e_m$ $e_t$ $e_v$ are fused through concatenation and passed to multi-layer perceptrons to make the final prediction $y_hat$. The framework of the entire model is shown in Figure \ref{approach2}.

\section{Experimental Setup}

\subsection{Dataset}
\begin{figure}[t]
\vskip 0.2in
\begin{center}
\centerline{\includegraphics[width=\columnwidth]{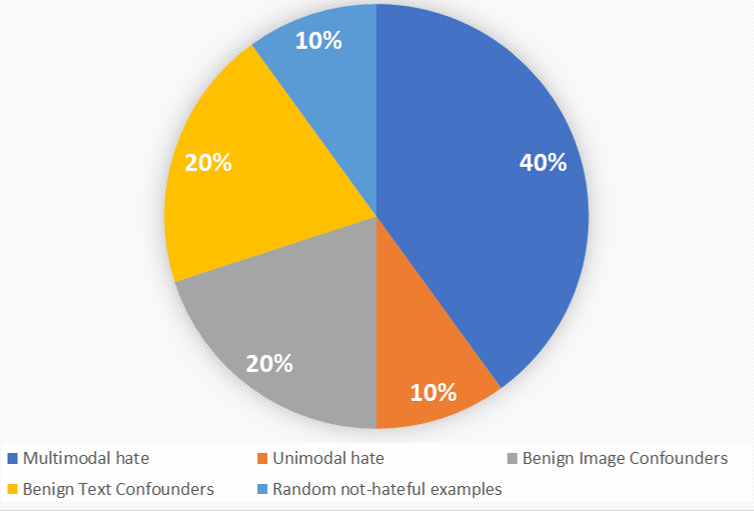}}
\caption{Types of memes in the Facebook Hateful Memes Challenge Dataset}
\label{type_of_memes}
\end{center}
\vskip -0.4in
\end{figure}

We have used the Facebook Memes Challenge Dataset \cite{kiela2020hateful} which comprises 10k memes. These memes are carefully designed for this task by annotators who are specially trained to employ hate-speech as defined by Facebook. The features in this dataset are the meme images themselves and string representations of the text in the image. The dataset comprises five different types of memes as shown in Figure \ref{type_of_memes}: multi-modal hate, where benign confounders were found for both modalities, unimodal hate where one or both modalities were already hateful on their own, benign image and benign text confounders and finally random not-hateful examples. The Training, Validation and Test split is 85, 5 and 10 respectively, and the individual sets are fully balanced. Each meme in the training and validation set are annotated as either 1 or 0 which corresponds to the hateful and benign classes respectively.

\subsection{Multi-modal Baselines}
\begin{figure}
\vskip 0.2in
\begin{center}
\centerline{\includegraphics[width=6cm, height =6cm]{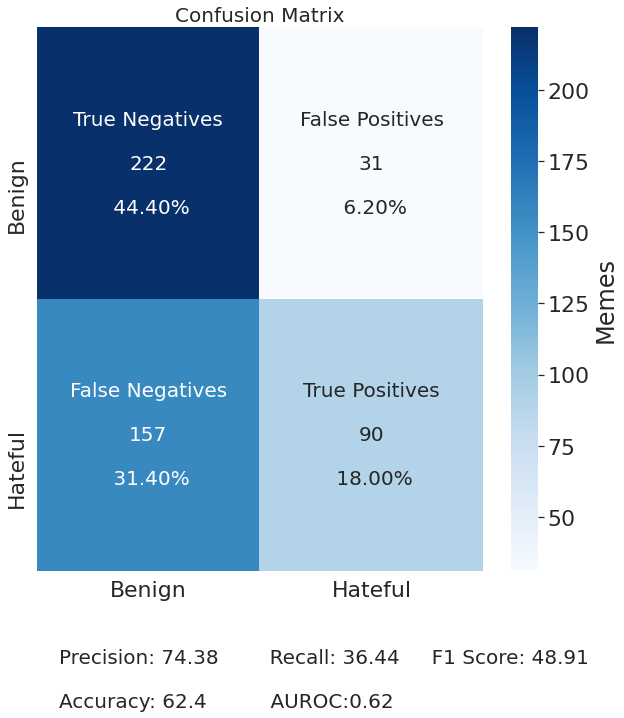}}
\caption{Confusion Matrix for baseline VisualBERT COCO model}
\label{Confusion Matrix for VisualBERT COCO model}
\end{center}
\vskip -0.4in
\end{figure}

For analysis, we select VisualBERT\cite{li2019visualbert}, a baseline model pretrained on COCO dataset with a multimodal objective. We fine-tune the model on our dataset following the same training guidelines as mentioned in the original challenge paper\cite{kiela2020hateful} and then evaluate it on the validation set comprising of 500 memes. Figure \ref{Confusion Matrix for VisualBERT COCO model} shows the Confusion matrix for the same, which gives an approximate of the error cases made by the baseline.

\subsubsection{VisualBERT}
In order to utilize the VisualBERT, multiple region features $f_1, f_2, …, f_n$ are first extracted from input image $I$ using Faster RCNN \cite{ren2015faster}. Each region feature $f$ is then converted to visual embedding $e_v$ by following equation.
\begin{equation}
e_v = f + e_s 
\end{equation}
where $e_s$ stands for segment embedding, which indicates whether the input is text or image. 

For the text input, the textual embedding $e_t$ is obtained in a similar way:
\begin{equation}
e_t = f_t + e_s + e_p
\end{equation}
where $f_t$ is the token embedding for each token in the sentence, and $e_p$ is the positional embedding to indicate the relative position of each token. After concatenating $e_v$ and $e_t$, the embedding is sent into pre-trained VisualBERT model for further processing. 

VisualBERT \cite{li2019visualbert} is a pre-trained model for learning joint contextualized representations of vision and language. It contains multiple transformer blocks on top of the visual and text embedding. It is pre-trained on Microsoft COCO captions \cite{chen2015microsoft} with two objectives: masked language modelling and sentence-image prediction task. The masked language modelling is very similar to the approach in sentence BERT \cite{devlin2018bert}, where some input text tokens are masked randomly, and the model needs to predict what are the original tokens. The sentence-image prediction requires the model to decide whether the input text matches the image. The VisualBERT output of the first token is used as the multi-modal representation $e_m$. An MLP is then used to make the final prediction. The model is fine-tuned for the current task by using the following loss function. 
\begin{equation} \label{crossentropy}
l(\theta) = CrossEntropyLoss(W \cdot e_m,y )
\end{equation}
where $e_m$ is a vector of size $h$. $h$ is the hidden size of VisualBERT. $W$, which has a shape of 2 by $h$, is the learnable matrix of the MLP. $\theta$ denotes the parameters of the entire model, including the $W$.

\subsection{Methodology}
For both approaches, we use mmf \cite{singh2020mmf}, a modular framework from Facebook AI Research, to build the main neural architectures. We use mmf's version of Visual BERT to generate multi-modal representations. The model is pre-trained on MS COCO dataset with a hidden dimension of 768.

For the Image Captioning models in our first approach, we use two implementations. The first one is an implementation of Show, Attend, and Tell by Xu et al. \cite{xu2015show} and second one using Bottom-Up Top-Down approach by Anderson et al. \cite{anderson2018bottom}. We take the top 10000 words from the vocabulary and process the images via Inception V3 model. The pre-trained Bert model used to encode generated caption has a dimension of 768. These two result are then fused together and then passed via an MLP classifier.

In the second approach, we directly use the final logits of sentiment analysis models and their sum as the sentiment embedding. The MLP classifer consists of 2 layers with 768 hidden units.

\subsection{Evaluation Metrics}
We have evaluated the performance of our classifier using the following two metrics as suggested in the challenge

\subsubsection{Area under the Receiver Operating Characteristics (AUCROC)}
Receiver Operating Characteristics curve is a graph of True Positive Rate (TPR) v/s False Positive Rate (FPR). It measures how well the binary classifier discriminates between the classes as its decision threshold is varied.\cite{quteprints114256}. A perfect classifier will have an area under the curve of 1, where the top left corner in the plot is the ideal point with a TPR of 1 and a FPR of 0. Thus, a larger area under the curve is desirable for any classifier to maximize TPR and minimize FPR.

\subsubsection{Classification Accuracy}
We find the accuracy of the predictions which is given by the ratio of correct predictions to the total number of predictions made, since it is easier to interpret. Thus, for each test sample, we output the label $\in \{0,1\}$ and the probabilities with which the classifier predicts a sample to be hateful. This probability is used to plot the AUCROC curve.

\begin{figure}[t]
\vskip 0.2in
\begin{center}
\centerline{\includegraphics[width=\columnwidth,height=6cm]{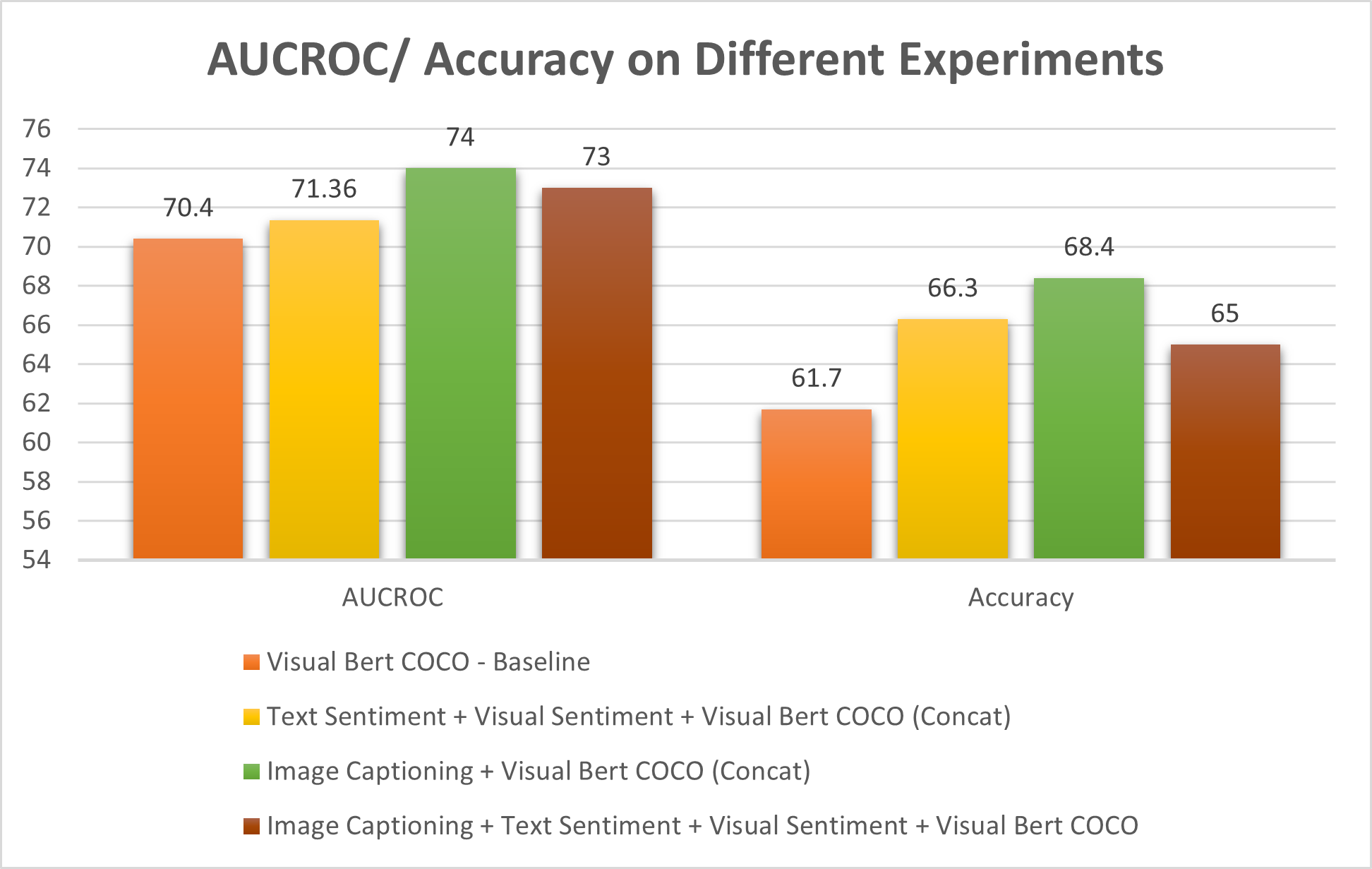}}
\caption{AUCROC/Accuracy for Different Experiments}
\label{results}
\end{center}
\vskip -0.4in
\end{figure}

\section{Results and Discussions}
\subsection{Image Captioning}
We use two frameworks to test our experiments, first being the MMF framework designed by the Facebook research lab who conducted this challenge and the second being creating all the models locally by using simple baselines like Concat BERT. 

Initially, we tested the image captioning locally by fusing it with Concat BERT baseline model . The baseline accuracy for this model turned out to be 57\%. Then, we built an Image captioning model based on Xu et al. \cite{xu2015show} and passed the caption via a Bert model to get the textual representation. When we fused this textual representation with the Concat BERT results, the accuracy increased by 2\% verifying the importance of the captioning and tackling the presence of the benign text confounders. Then, we shifted to the MMF framework to test it on better baseline models like Visual Bert. 
As can be seen in the Figure \ref{results}, the image captioning approach gives a significant improvement in the AUCROC and the accuracy on the test set. There is increase of 3.6 \% in the AUCROC score and an increase of 6.7 \%  in the accuracy of the model. This shows us that the image captioning model tackles these benign confounders and give a better representation to the image modality and thus improves the results.

\begin{figure}
\vskip 0.2in
\begin{center}
\centerline{\includegraphics[width=\columnwidth,height= 5 cm]{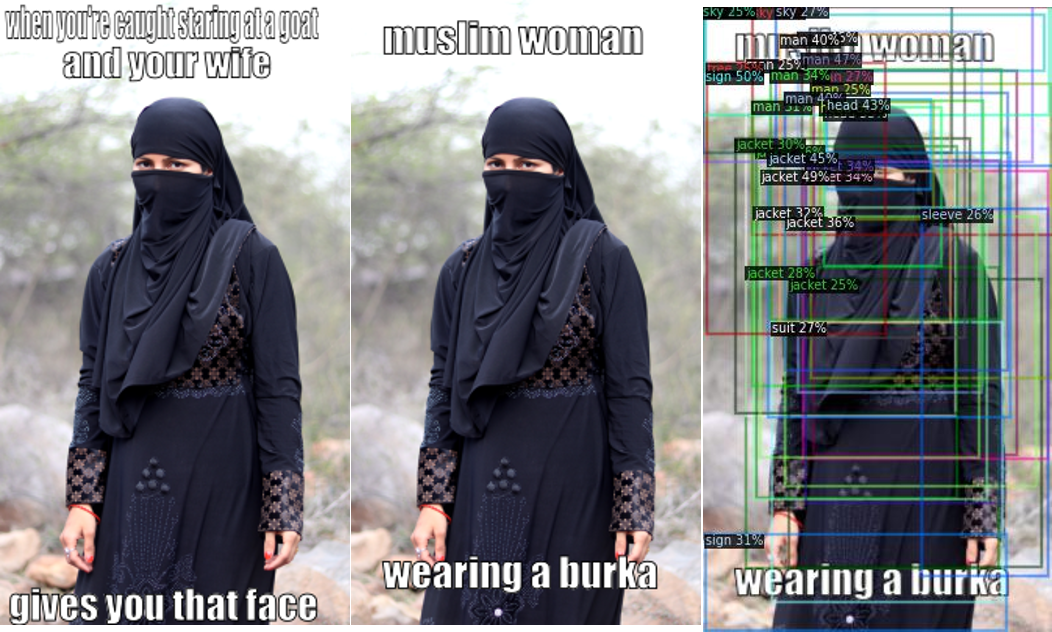}}
\caption{Mean meme (left), Benign Text Confounder and the testing meme (middle), Object Detection Visualization before captioning(right)}
\label{Image Captioning Example}
\end{center}
\vskip -0.4in
\end{figure}

The Figure \ref{Image Captioning Example} comprises of three images. The first image is the original hateful meme, the second image is the one being tested which is created by adding benign text confounder by just describing the image and thus making it a non-hateful meme with a label of '0'. i.e., non-hateful. The third image shows the visualization of object detection bounding boxes on the test image. For the test image as input, the baseline VisualBert  predicts a label '1', thus, misclassifying it as a hateful meme because it is not able to understand the benign text confounder. However, using our approach, it is correctly labeled as benign. Our model captions the image similar to the benign text confounder and thus the model learns about its similarity and that its benign behavior. This helps the classifier to classify this as a benign result. There are many such examples present in the dataset which are correctly classified by our model, thus improving the accuracy and the AUCROC value. We also ran Bilinear Transformation as the fusion technique but it brought the performance down and also it ran very slow on the dataset, thus, we decided to move ahead with concatenation itself for the results. 

\begin{figure}[t]
\vskip 0.2in
\begin{center}
\centerline{\includegraphics[width=\columnwidth,height= 5 cm]{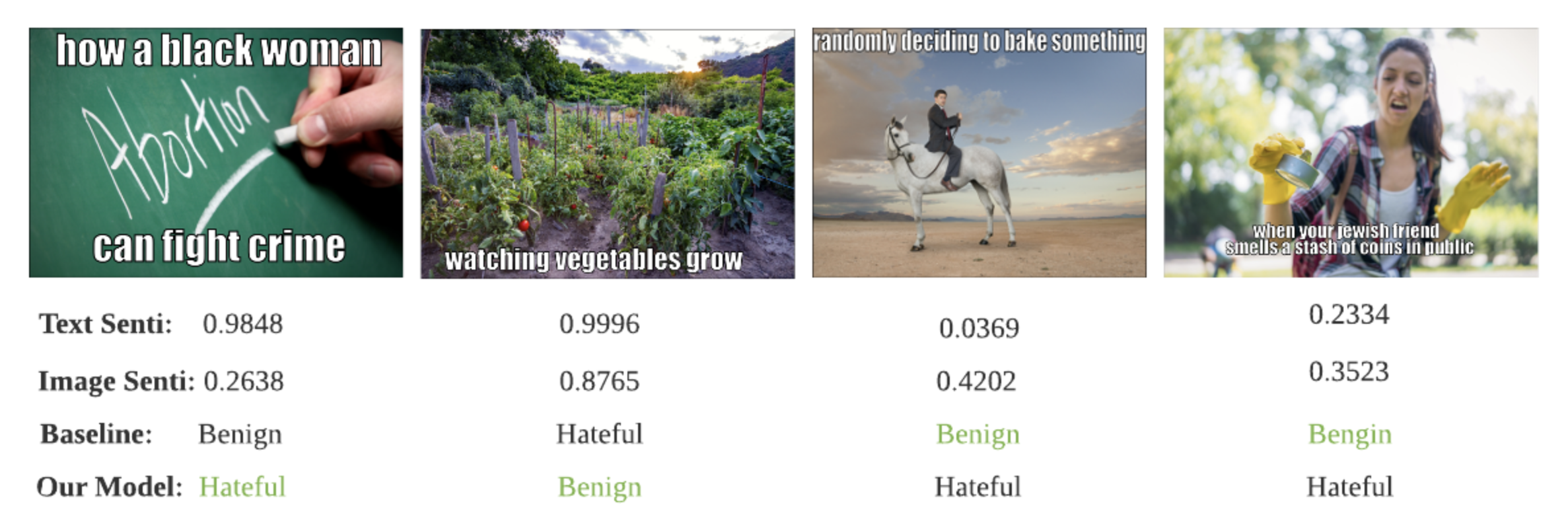}}
\caption{Sample images from dev set. The sentiment value under the image are ranged from 0 to 1 with 1 as positive. Green label denotes the ground truth. }
\label{senti_res}
\end{center}
\vskip -0.4in
\end{figure}

\subsection{Sentiment Analysis}
For the sentiment analysis approach, although the model doesn't improve the AUCROC value by a large margin, we still see a significant gain of 4$\%$ in the accuracy. We directly compare our models' results against the Visual BERT baseline and observe two common cases when sentiment analysis benefits the prediction. The first case is when the text and image have opposite sentiments, as shown in the first image of Figure \ref{senti_res}. The baseline considers this meme as benign, but our model can clearly indicate its irony and then guide the prediction. The other is when both modalities have a positive sentiment as the meme shown in the second image of Figure \ref{senti_res}. Sentiment information can help to confirm benign memes. However, since we do not have the annotated data to fine-tune the sentiment analysis models or perform multi-task learning, the accuracy of sentiment prediction is undesirable. As shown in the third meme in Figure \ref{senti_res}, the text doesn't seem very negative, and the image seems neutral, but our model predicts both as negative. Also in some complicated cases, the sentiments are not very helpful. For example, when sentiments of both modalities are negative as the last two visuals in the figure, our model does not work well because the meme has a similar chance to be benign or hateful.

\subsection{Combining both Image Captioning and Sentiment Analysis}

\begin{figure}
\vskip 0.2in
\begin{center}
\centerline{\includegraphics[width=\columnwidth,height= 5 cm]{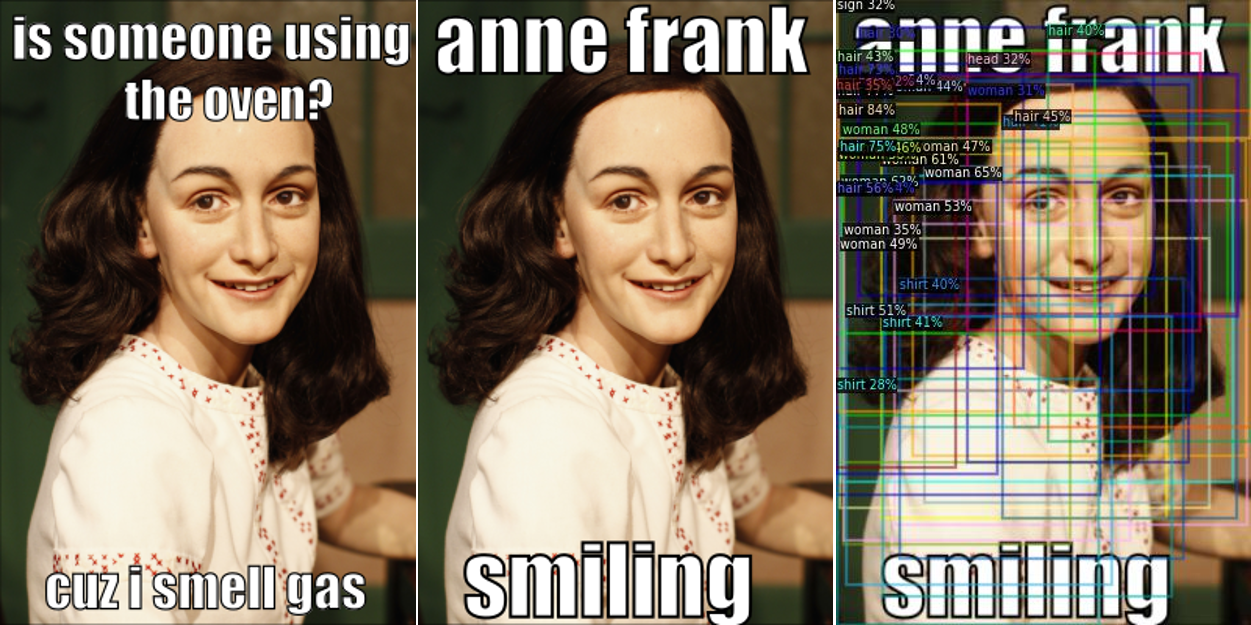}}
\caption{Mean meme (left), Benign Text Confounder and the testing meme with positive text sentiment and positive visual sentiment (middle), Object Detection Visualization before captioning(right)}
\label{Combined Example}
\end{center}
\vskip -0.4in
\end{figure}

We also performed an experiment wherein we concatenated both the image captioning results as well as the sentiment analysis features along with the Visual Bert multimodal representation and fine tuned it on the dataset. Again, we saw a significant increase in the AUCROC and the accuracy of the model in comparison to the baseline model. We expected the results to give an even better performance than the captioning results, as it would have different features to learn from, but the value of the accuracy decreased in comparison to the Image captioning results. Some reasons for this behavior could be due to conflicts in the two representations being concatenated together which could lead to lower accuracy and AUCROC value. Another reason could be due to presence of redundant features in different representations and thus reducing the performance. 
We also performed an analysis on some data points related to this test. As can be seen in the figure \ref{Combined Example}, the middle image of the benign confounder is wrongly classified by the baseline model as hateful but the combined approach learns the alignment of the caption and the pre-extracted caption along with the sentiment of both the modalities (i.e. positive in this case) and gives a correct prediction of non-hateful label.

\section{Conclusion and Future Directions}
We present two novel mediums for introducing outside world knowledge to our multi-modal models. i.e. Image Captioning and Sentiment analysis. Our approach enables to identify the adversarial examples in the Hateful Memes Challenge dataset. While both Image Captioning and Sentiment analysis show a promising improvement over the baseline models published by the Facebook challenge, the combination of object detection and image captioning provides the best results.

One of the primary objectives of this challenge was to facilitate the research for the development of true multi-modal models which gives importance to all the modalities. After analysis of the dataset and various techniques for this task, we have identified several areas which should be explored for future research in this domain. An improvement in the quality of captions generated by other image captioning models like OSCAR \cite{li2020oscar}  will enhance the ability of the model to identify benign text confounders and thus increase the classification accuracy. Combining image captioning and sentiment in an efficient manner such that they cancel out their individual effects is critical. Fusion plays a key role in this task, thus we plan to explore more concatenation techniques, by using attention mechanism, transformers etc.  Further, we believe that using other Large-scale pretrained multi-modal models like UNITER \cite{chen2020uniter}, providing ‘Internet Knowledge’ through graphical approach are some interesting research questions in this task.

\FloatBarrier
\bibliography{example_paper}
\bibliographystyle{icml2020}

\end{document}